\documentclass[authoryear,preprint,12pt]{elsarticle}
\usepackage{color}
\pdfoutput=1
\usepackage{amsfonts,graphicx,subfigure,algorithm,algorithmicx}
\usepackage[noend]{algpseudocode}
\usepackage{xcolor}
\usepackage{multirow}
\usepackage{makecell}
\usepackage{booktabs}  
\usepackage{float}
\usepackage{wrapfig}

\usepackage{amssymb}
\usepackage{amsmath}

\journal{Applied Soft Computing}

\begin{document}

\begin{frontmatter}



\title{AdaSin: Enhancing Hard Sample Metrics with Dual Adaptive Penalty for Face Recognition}


\author[1]{Qiqi Guo} 
\ead{chet77@stu.jnu.edu.cn}

\author[1]{Zhuowen Zheng}
\ead{202334261141zzw@stu.jnu.edu.cn}

\author[1]{Guanghua Yang}
\ead{ghyang@jnu.edu.cn}

\author[2]{Zhiquan Liu}
\ead{zqliu@vip.qq.com}

\author[1]{Xiaofan Li}
\ead{lixiaofan@jnu.edu.cn}

\author[3]{Jianqing Li}
\ead{jqli@must.edu.mo}

\author[3]{Jinyu Tian}
\ead{jytian@must.edu.mo}

\author[1]{Xueyuan Gong \corref{cor1}}
\ead{xygong@jnu.edu.cn}

\address[1]{School of Intelligent Systems Science and Engineering, Jinan University, Guangdong, China}
\address[2]{College of Cyber Security, Jinan University, Guangdong, China}
\address[3]{School of Computer Science and Engineering, Macau University of Science and Technology, Macau, China}

\cortext[cor1]{Corresponding author}

\begin{abstract}
In recent years, the emergence of deep convolutional neural networks has positioned face recognition as a prominent research focus in computer vision. Traditional loss functions, such as margin-based, hard-sample mining-based, and hybrid approaches, have achieved notable performance improvements, with some leveraging curriculum learning to optimize training. However, these methods often fall short in effectively quantifying the difficulty of hard samples. To address this, we propose Adaptive Sine (AdaSin) loss function, which introduces the sine of the angle between a sample's embedding feature and its ground-truth class center as a novel difficulty metric. This metric enables precise and effective penalization of hard samples. By incorporating curriculum learning, the model dynamically adjusts classification boundaries across different training stages. Unlike previous adaptive-margin loss functions, AdaSin introduce a dual adaptive penalty, applied to both the positive and negative cosine similarities of hard samples. This design imposes stronger constraints, enhancing intra-class compactness and inter-class separability. The combination of the dual adaptive penalty and curriculum learning is guided by a well-designed difficulty metric. It enables the model to focus more effectively on hard samples in later training stages, and lead to the extraction of highly discriminative face features. Extensive experiments across eight benchmarks demonstrate that AdaSin achieves superior accuracy compared to other state-of-the-art methods.
\end{abstract}



\end{frontmatter}



\section{Introduction}
\label{sc:intro}

In recent years, Deep Convolutional Neural Networks (DCNNs) have significantly advanced the field of computer vision, with face recognition emerging as a prominent task. Designing an effective loss function is crucial for this endeavor.

Previous studies \citep{liu2017sphereface}, \citep{wang2018cosface}, and \citep{deng2019arcface}, have demonstrated that margin-based loss functions address the limitations of traditional softmax loss in extracting highly discriminative embedding features. These loss functions enhance inter-class variation and intra-class compactness to a certain degree. However, they fail to distinguish between easy and hard sample regions, which restricts the model's capacity to learn features with higher discriminative power. To tackle this issue, several works have introduced hard sample mining techniques. For instance, \citep{shrivastava2016training} empirically identifies and retains only a specific proportion of hard samples, discarding easy ones entirely. Similarly, \citep{ross2017focal} employs multiple hyperparameters to reduce the weight of easy samples, thereby focusing more on hard samples. In essence, mining-based loss functions aim to amplify the impact of hard samples, as highlighted by \citep{schroff2015facenet}. More recently, MV-Softmax \citep{wang2020mis} combined mining-based and margin-based loss functions, explicitly identifying misclassified samples as hard samples. MV-Softmax proposes that the cosine similarity of hard samples to other negative class weight centers (i.e., negative cosine similarity), can be used to learn highly discriminative features. From a probabilistic perspective, \citep{wang2020mis} uses a multiplier greater than 1 to amplify the negative cosine similarity of hard samples, thereby reducing the probability of hard sample vectors, which can increase the model's focus on hard samples.

\begin{figure}[!t]
	\centering
        \includegraphics[width=0.7\linewidth]{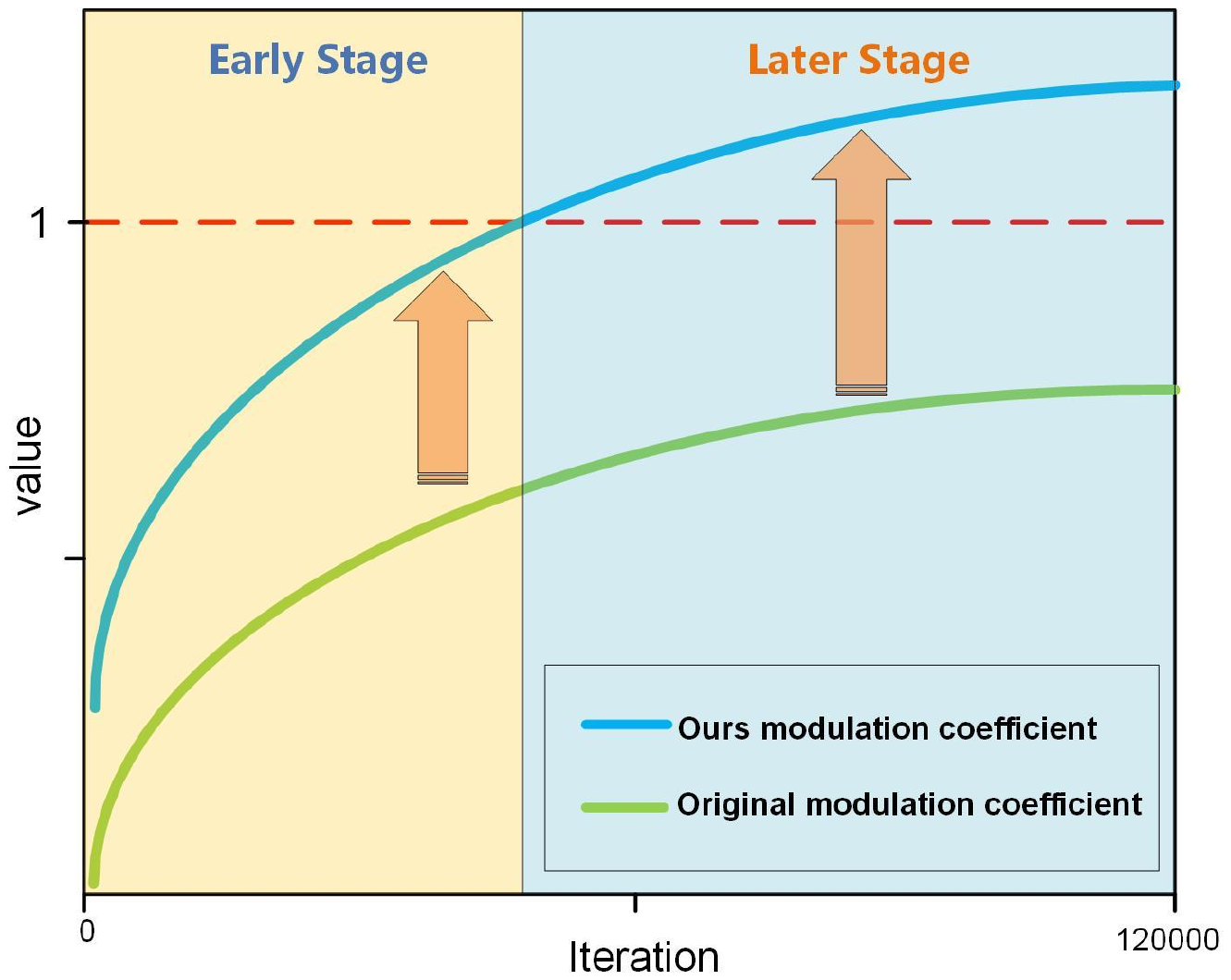}
	\caption{Comparison of the changes in the values of the modulation coefficient function of our design with the values of the original modulation coefficient function.}
	\label{fg:mc1}
\end{figure}

The concept of curriculum learning was first introduced by \citep{bengio2009curriculum}, while \citep{huang2020curricularface} applied this idea to face recognition and proposed a novel loss function called CurricularFace. This method introduces an adaptive modulation coefficient function $I(t, \cos{\theta_j})$, which increases as training progresses, adjusting the weighting of the negative cosine similarity for hard samples. In the early training stages, the model focuses on learning from easy samples, gradually shifting its attention to hard samples as the training progresses. This progressive learning strategy allows the model to establish a solid foundation by learning discriminative features from easy examples, followed by refining its understanding through more challenging hard samples. However, we found that CurricularFace relies on the negative cosine similarity to measure the difficulty of the samples and uses it to calculate the modulation coefficient to be not optimal. The modulation coefficient $I(t, \cos{\theta_j})$ calculated using such an approach is unlikely to exceed 1 (see the green curve in Fig. \ref{fg:mc1}), and does not serve to amplify the negative cosine similarity. In the later stages of training, using this modulation coefficient function to reweight the negative cosine similarity of hard samples instead reduces the model's focus on these hard samples. Moreover, both CurricularFace and MV-Arc-Softmax do not take into account the adaptivity of the cosine similarity between the sample and the positive class center (i.e., the positive cosine similarity).

Furthermore, since each sample corresponds to multiple negative class centers while the positive class center is unique, the use of negative cosine similarity to measure the difficulty of hard samples makes it difficult to adaptively adjust the positive cosine similarity for hard samples. Therefore, we propose to measure the difficulty of a sample solely based on the sine of the angle between the sample and the positive class center. This way, we only need one measurement to simultaneously apply adaptive penalties to both the positive and negative cosine similarities of the sample. We refer to the method that adaptively adjusts both positive and negative cosine similarities as the dual adaptive penalty.

In this paper, we propose a novel loss function that improves the measurement of hard sample difficulty by leveraging an adaptive curriculum learning strategy and integrating an adaptive margin loss function. Specifically, we define the difficulty measure for hard samples as $D(\theta_k) = \sin\left(\frac{\theta_k}{2}\right)$, where $\theta_k \in [0, \pi]$ denotes the angle between the $k$-th sample and the positive class center. Intuitively, larger angles correspond to greater difficulty, resulting in higher values of $D(\theta_k)$. To incorporate this measure, we design an improved modulation coefficient function $\Phi(t,D(\theta_k))$. Our modulation coefficient not only reweights the negative cosine similarity of samples but also adaptively adjusts the angular margin in the positive cosine similarity of samples. The proposed loss function combines adaptive margins with a hard sample mining strategy, offering two key advantages: (1) Since the difficulty measurement function for samples is non-negative within its domain, this ensures that the value of the modulation coefficient function will exceed 1 in the later stages of training (See the blue curve in Fig. \ref{fg:mc1}), effectively emphasizing hard samples. (2) We have redesigned the modulation coefficient function to suit the dual adaptive penalty. In simpler terms, while it amplifies the negative cosine similarity of hard samples, it also adaptively adjusts the angular margin to reduce the positive cosine similarity of the sample. This allows the model to more effectively emphasize hard samples.

In summary, our approach provides a stronger metric for quantifying the hardness of a sample. By incorporating dual adaptive penalty, our method enhances the penalization of hard samples, enabling the model to learn more discriminative features. Experimental results show that the sinusoidal-based dual adaptive penalty method is particularly effective in managing hard samples, leading to significant improvements in model accuracy. The main contributions of this paper are summarized as follows:

\begin{itemize}
	\item To our knowledge, we are the first to propose using the sine of the angle between the sample and the positive class center as a measure of hard sample difficulty, and improved the previous modulation coefficient function.
	\item We design a novel loss function that can adaptively adjust both positive cosine similarity and negative cosine similarity.
	\item We conducted extensive experiments on the popular benchmark LFW \citep{huang2008labeled}, CALFW\citep{zheng2017cross}, CPLFW\citep{zheng2018cross}, AgeDB-30\citep{moschoglou2017agedb}, CFP-FP\citep{sengupta2016frontal}, VGG2-FP\citep{cao2018vggface2}, IJBB\citep{whitelam2017iarpa}, and IJBC\citep{maze2018iarpa}, which proved the superior of Adasin over baseline and other state-of-the-art (SOTA) competitors.
\end{itemize}

\section{Related Work}
\label{sc:rw}
\subsection{Margin-based and mining-based loss function}

Margin-based loss functions have marked a significant milestone in advancing face recognition tasks. Most existing margin-based loss functions are derived from the foundational softmax classification loss \citep{taigman2014deepface}. However, the traditional softmax loss function is limited in its ability to enable the model to learn highly discriminative features. To address this limitation, various margin-based loss functions, such as \citep{liu2017sphereface}, \citep{wang2018cosface}, and \citep{deng2019arcface}, have been introduced. Building on these developments, adaptive margin loss functions, such as \citep{boutros2022elasticface}, \citep{jiao2021dyn},  \citep{li2024hamface}, \citep{xu2024x2} and \citep{kim2022adaface}, have emerged and moving beyond fixed-margin designs. While these adaptive approaches improve model accuracy, they generally treat all samples uniformly and fail to distinguish between easy and hard samples. The margin function $f$ can be expressed in a general form as follows: $f (m, \theta_{y_i}) = \cos{(m_1\theta_{y_i} + m_2) - m_3}$. where $m_1$, $m_2$, and $m_3$ can either be fixed hyperparameters or adaptive variables. Similar loss functions are not limited to face recognition, \citep{alirezazadeh2022deep} designed a loss function for fashion clothing classification based on additive cosine margins.

Mining-based loss functions, such as Online Hard Example Mining (OHEM) \citep{shrivastava2016training} and Focal Loss \citep{ross2017focal}, are rarely employed in face recognition tasks. OHEM is a dynamic sample selection method designed to enhance model performance by prioritizing hard samples. It selects the most challenging samples within each training batch, which are the ones the model finds hardest to classify correctly, thereby improving the model's ability to learn from these difficult instances. Focal Loss, on the other hand, extends the standard cross-entropy loss by introducing a balancing factor, $\alpha_t$, and a modulating factor, $\gamma$. These factors control the emphasis placed on hard samples, assigning them higher weights in the loss calculation. This mechanism increases the influence of hard samples on the training process, enabling the model to focus more effectively on these challenging cases. There are also studies \citep{gao2020multi} on multi-scale patch representation feature learning schemes to solve the low-resolution face recognition problem.

\subsection{MV-Softmax and CurricularFace}

MV-Softmax \citep{wang2020mis} is specifically designed to address the challenge of mining hard samples in face recognition. It is the first loss function to integrate mining-based and margin-based strategies into a unified framework, as well as the first to explicitly define misclassified samples as hard samples. The hyperparameter $t$ is introduced to amplify the weight of the negative cosine similarity for hard samples. By penalizing these hard samples more heavily, MV-Softmax effectively drives the model to focus on learning from challenging instances, thereby enhancing its ability to handle hard sample.

Curriculum learning (CL) was first introduced by \citep{bengio2009curriculum}, drawing inspiration from the human learning process, which begins with easy tasks before gradually transitioning to more challenging ones. In the judgment of hard samples, CurricularFace follows the approach of MV-Softmax. The difference is that CurricularFace modifies the original hyperparameter $t$ to an adaptive parameter. A key component of CurricularFace is its method for measuring the difficulty of hard samples. It uses the negative cosine similarity of the samples, where a smaller angle between the sample and the negative class indicates higher difficulty. To implement this, a modulation coefficient function $I(t, \cos{\theta_j})$ is introduced, where $t$ is an adaptive parameter that increases as training progresses.

In contrast, our method employs a sinusoidal relationship to measure the difficulty of hard samples, using the angle between the sample and the positive class center. This alternative metric provides a more nuanced and effective approach to evaluating and penalizing hard samples.

\section{Proposed Approaches}
\label{sc:pa}

\subsection{Traditional softmax-based loss function and its variants}

The traditional softmax-based loss function can be formulated as:
\begin{equation}
L_1 = -\frac{1}{B}\sum_{i=1}^{B}\log\frac{e^{W_{y_i}^T x_i + b_{y_i}}}{\sum_{j=1}^{n}e^{W_j^T x_i+b_j}},
\label{eq:tsl}
\end{equation}
where $x_i \in \mathbb{R}^{d} $ represents the feature vector of the $i$-th sample, which belongs to the class $y_i$, $ W_j\in\mathbb{R}^{d}$ denotes the $j$-th column of weights $W\in\mathbb{R}^{d \times n}$, which also denotes the $j$-th class, and $b_j$ is the bias term. The dimension of the embedding features is $d$ and the number of classes is $n$. The batch size is $B$. Following \citep{wang2017normface,liu2017sphereface,wang2018cosface,ranjan2017l2}, we let each weight be set to $||W_j||=1$ by $l_{2}$ normalisation, with the embedding feature $||x||$ fixed to $s$ and the bias value $b_j=0$. So the logits translate into $ W_j^T x_i =s\cos{\theta_j}$, where $\theta_j$ is the angle between the weight $W_j$ and the embedding feature $x_i$. Thus the traditional softmax-based loss function can be modified as:

\begin{equation}
    L_2 = -\frac{1}{B}\sum_{i=1}^{B}\log\frac{e^{s(\cos\theta_{y_i})}}{e^{s(\cos \theta_{y_i})}+\sum_{j=1, j\neq y_i}^{n}e^{s(\cos\theta_j)}}
\label{eq:mtsl}
\end{equation}
As seen from Eq. \ref{eq:mtsl}, the modified traditional softmax-based loss function optimizes the model by relying solely on the angle between the feature vectors and the class center weights, which facilitates faster convergence. However, after this modification, the decision boundary still corresponds to the angle bisector between the two classes, and the range of variation for samples in the angular space remains large. This is not conducive to the optimization of the model. To address these limitations, many variants have been proposed in previous studies. These traditional Softmax-based loss functions can generally be formulated as:

\begin{equation}
    L_3 = -\frac{1}{B}\sum_{i=1}^{B}\log\frac{e^{sT(\theta_{y_i}, m)}}{e^{sT(\theta_{y_i},m)}+\sum_{j=1, j\neq y_i}^{n} e^{sN(\theta_j)}},
    \label{eq:gtsl}
\end{equation}
The primary distinction among these variants lies in the design of the positive cosine similarity $T(\cos\theta_{y_i})$ and negative cosine similarity $N(\theta_j)$. In the margin-based loss function, ArcFace \citep{deng2019arcface} designs the positive cosine similarity as $T(\theta_{y_i}, m) = \cos({\theta_{y_i}} + m)$ for rach sample. CosFace\citep{wang2018cosface} and SphereFace\citep{liu2017sphereface}, on the other hand, use cosine margin $T (\theta_{y_i}, m)=\cos({\theta_{y_i}})-m$ and multiplicative margin $T (\theta_{y_i}, m)=\cos(m{\theta_{y_i}})$, respectively. Their negative cosine similarities are the same as in Eq. \ref{eq:gtsl}, i.e., $N (\theta_j)=\cos{\theta_j}$. All of these methods modify only the positive cosine similarity. Their decision boundary can be expressed as: $T(\theta_{y_i}, m)=N( \theta_j)$. A recent study, MV-arc-softmax \citep{wang2020mis}, suggests that the negative cosine similarity of samples can also play a crucial role in learning highly discriminative features. Therefore, the modified negative cosine similarity $N(t, \theta_j)$ can be formulated as follows:

\begin{equation}
    N(t, \theta_j) = 
    \begin{cases} 
        \cos\theta_j, &T(\theta_{y_i}, m)\geq\cos\theta_j   \\ 
         (t+1)\cos{\theta_j} + t, & T(\theta_{y_i}, m)<\cos\theta_j.
    \end{cases}
    \label{eq:MV}
\end{equation}
where $t$ is an artificially set hyperparameter, and in the case of MV-arc-softmax generally $t>0$. If a sample is recognized as a hard sample, its negative cosine similarity is scale up to $(t+1)\cos{\theta_j}+t$. otherwise, it remains unchanged.

\begin{figure}[H]
	\centering
	\subfigure[The negative cosine similarity]{
		\includegraphics[width=0.45\columnwidth]{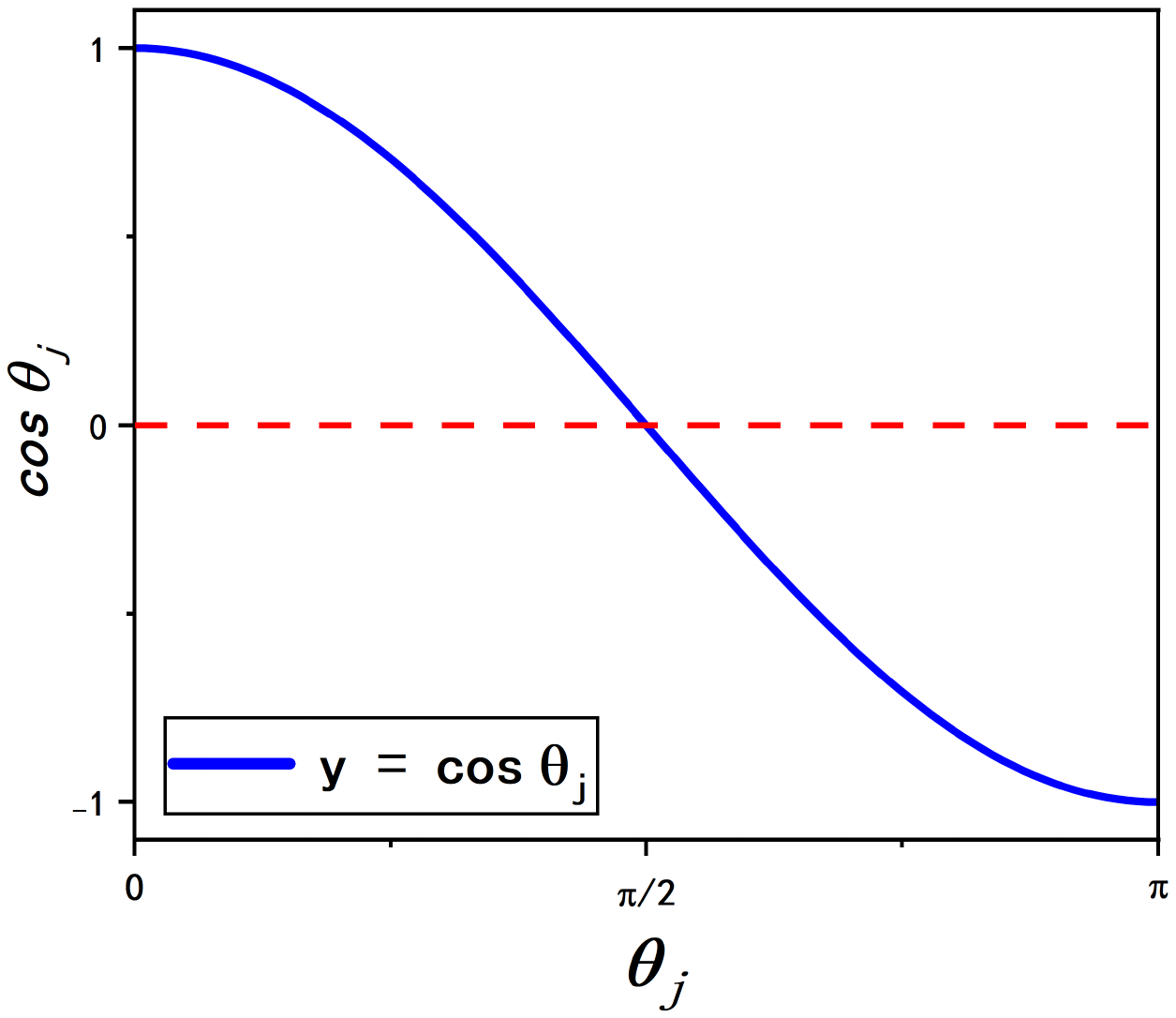}
	}
	\hfil
	\subfigure[We designed the metric function]{
		\includegraphics[width=0.45\columnwidth]{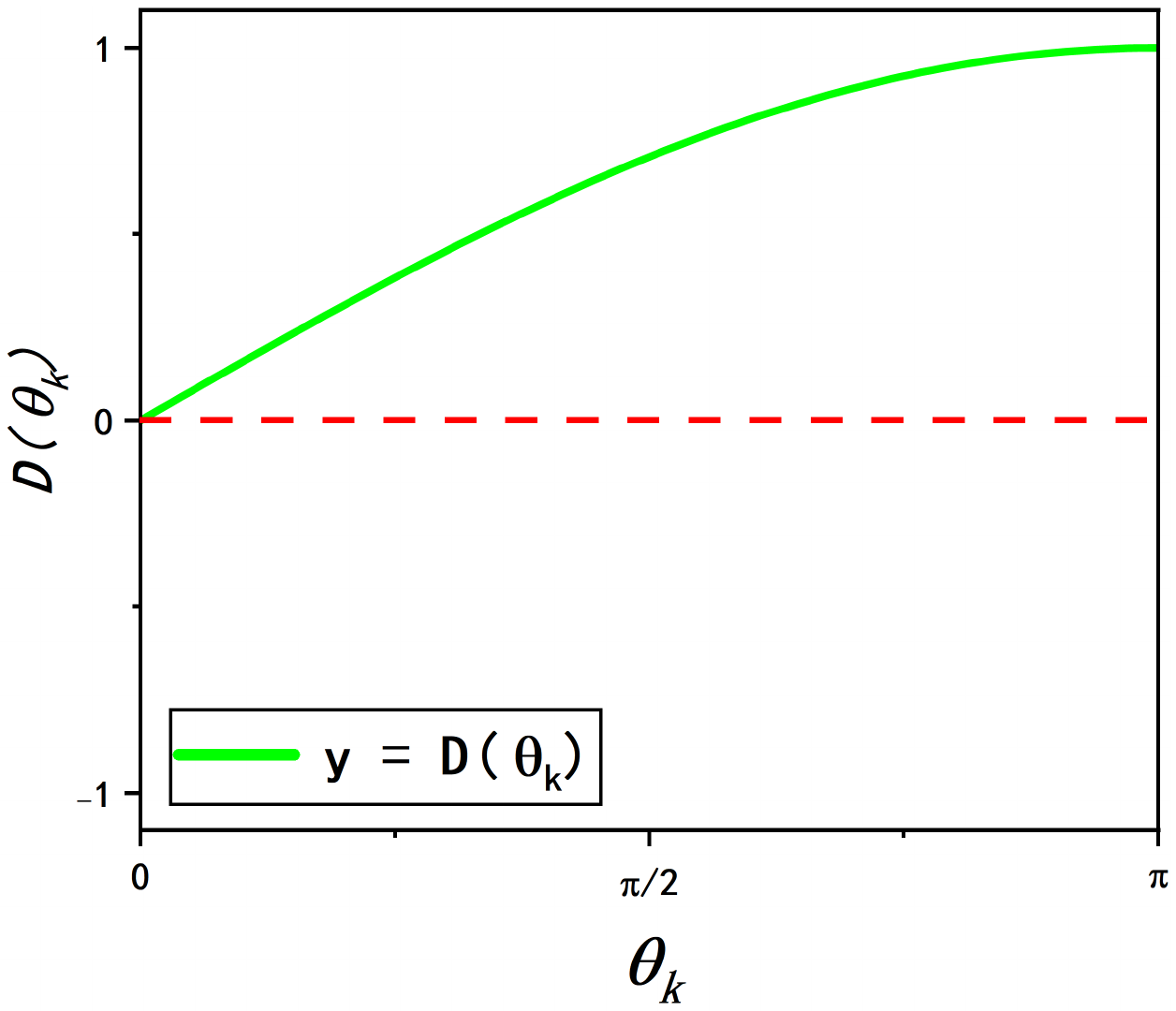}
	}
	\caption{The range of values of the two metric functions.}
	\label{fg:function}
\end{figure}

CurricularFace believes that at the early training stage, the model should emphasize easy samples first, and then focus hard samples at a later stage. To achieve this, the parameter $t$ is made adaptive. Specifically, $t$ is set to the average of the positive cosine similarity of a batch. To avoid instability caused by extreme data points in a batch, an Exponential Moving Average (EMA) is used to update $t$. Its expression can be formulated as follows:

\begin{equation}
    t^{(k)}=\alpha r^{(k)}+(1-\alpha)t^{(k-1)},
    \label{eq:ema}
\end{equation}
where $t^{(0)}=0$ and $\alpha$ is the momentum parameter set to 0.99. $r^{(k)} = \frac{1}{B}\sum_{i}^{B} \cos{\theta^{(k)}_{y_i}}$ denotes the average positive cosine similarity of the $k$-th batch. With these settings, CurricularFace is designed with a modulation coefficient function $I(t^{(k)}, \cos{\theta_j})= t^{(k)}+\cos{\theta_j}$, which adapts well to the training phase. The negative cosine similarity is improved to: 

\begin{equation}
    N(t^{(k)}, \theta_j) = 
    \begin{cases} 
        \cos \theta_j, &T(\theta_{y_i}, m)\geq\cos{\theta_j}   \\ 
         I(t^{(k)},\cos{\theta_j})\cos{\theta_j}, &T(\theta_{y_i}, m)<\cos{\theta_j}.
    \end{cases}
    \label{eq:curricularface}
\end{equation}
The modulation coefficient increases as training progresses, and within the same training stage, more hard samples are assigned a larger modulation coefficient. Ideally, after a certain point in the training process, the modulation coefficient $I(t^{(k)}, \cos{\theta_j})>1$, which amplifies the negative cosine similarity of hard samples. In this method, negative cosine similarity is used to measure the difficulty of hard samples. However, the values of negative cosine similarity are typically quite small and can even be negative, as shown in Fig. \ref{fg:function}a. As a result, the modulation coefficient function $I(t^{(k)}, \cos{\theta_j})$ is unlikely to exceed 1 during the entire training process, as demonstrated by the green curve in Fig. \ref{fg:mc2}. Additionally, since each sample has multiple negative cosine similarities, it means that each sample can have multiple modulation coefficient functions. Although this easily allows for reweighting the negative cosine similarity, it is not suitable for the margin adaptivity in the positive cosine similarity.

\subsection{Adaptive Sine Loss}

\begin{figure*}[!t] 
\centering 
\includegraphics[scale=0.49]{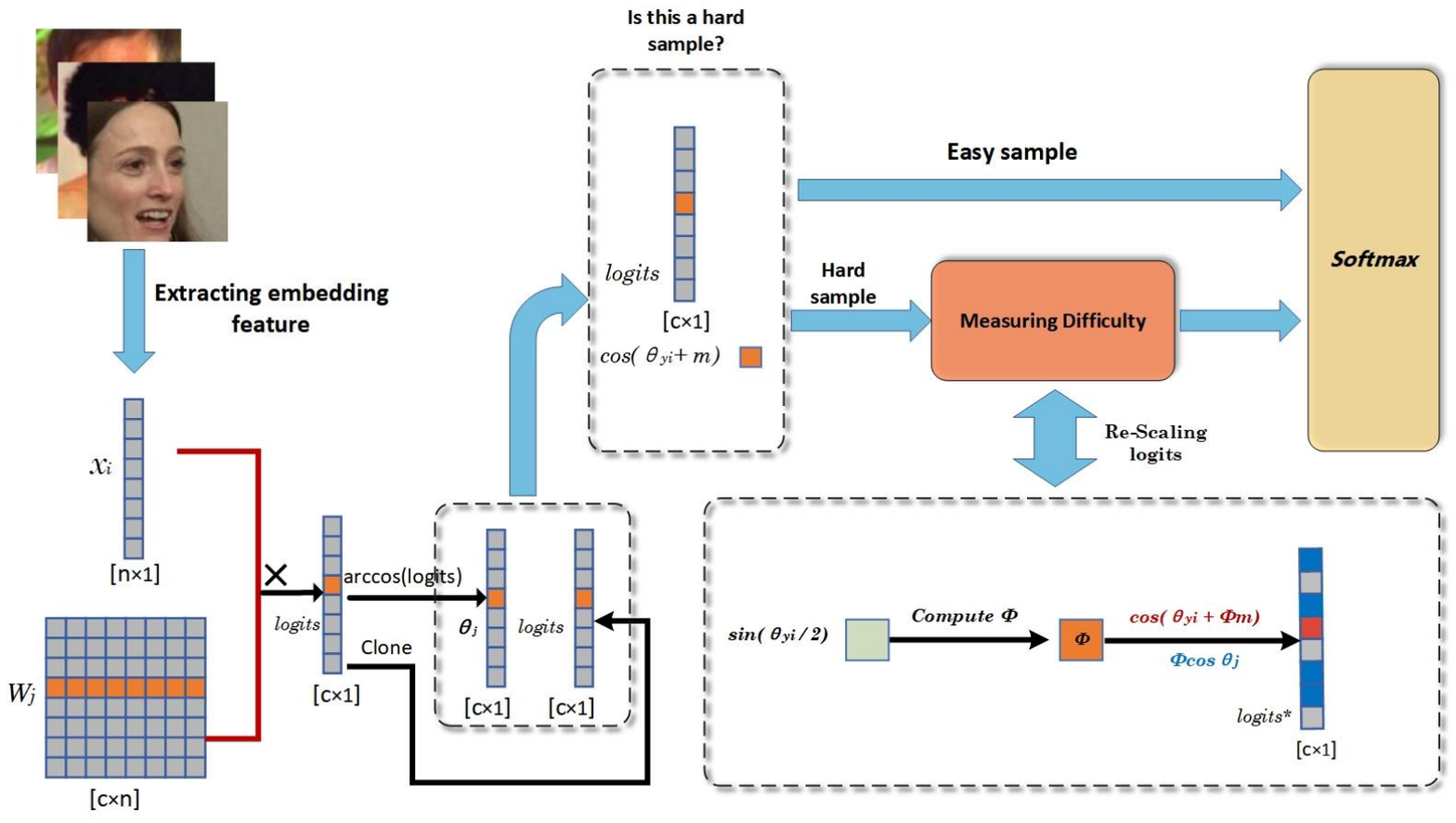} 
\caption{The figure above illustrates the specific workflow of our AdaSin loss function. First, we determine whether a sample is a hard sample. If the sample is classified as easy, we directly proceed with the softmax computation. If it is identified as a hard sample, we calculate the sine value of the angle between the sample and the positive class center. Next, we compute the modulation coefficient $\Phi$, and finally, we apply this coefficient to adaptively scale the logits.} 
\label{fg:flow} 
\end{figure*}
Next, we present our proposed dual adaptive penalty method based on the sine metric. We modify the original difficulty measurement approach by replacing the cosine of the angle between the sample and positive class center with the sine of the angle. This modification allows us to implement dual adaptive penalty for both positive and negative cosine similarities. Our loss function is also expressed in the general form of Eq. \ref{eq:gtsl}, with positive and negative cosine similarities defined as follows:

\begin{equation}
T( \theta_{y_i},m,\Phi)  = \\
    \begin{cases} 
        \cos{(\theta_{y_i}+m)}, & \cos{(\theta_{y_i}+m)}\geq\underset{0\leq j\leq n-1}{\max}\cos\theta_j  \\ 
        \cos{(\theta_{y_i}+\Phi\ast m)}, &\cos{(\theta_{y_i}+m)}<\underset{0\leq j\leq n-1}{\max}\cos\theta_j,
    \end{cases}
\label{eq:myT}
\end{equation}

\begin{equation}
    N( \Phi, \theta_j) = 
    \begin{cases} 
        \cos \theta_j, &\cos{(\theta_{y_i}+m)}\geq\cos\theta_j  \\ 
        \Phi\ast\cos\theta_j, &\cos{(\theta_{y_i}+m)}<\cos\theta_j.
    \end{cases}
\end{equation}
\label{eq:myN}The positive cosine similarity we define can be formulated in the same way as any margin-based loss function, and here we use ArcFace as an example. In Eq. \ref{eq:myT}, $\underset{0\leq j\leq n-1}{\max}\cos\theta_j$ denotes the maximum negative cosine similarity of the samples. The modulation coefficient function we designed can be formulated as follows:

\begin{equation}    
     \Phi= t^{(k)}+h\ast D(\theta_{y_i}),
\label{eq:scale}
\end{equation}
where the adaptive parameter $t^{(k)}$ follows the calculation of CurricularFace\citep{huang2020curricularface} i.e. the average of the positive cosine similarity of the $k$-th batch, as shown in Eq. \ref{eq:ema}. It is worth mentioning that we multiply a hyperparameter $h$ with the metric function $D(\theta_k)$. The purpose of this is to prevent the modulation coefficient from becoming too large. It also helps control the timing of when the model starts emphasizing hard samples, preventing the model from focusing too early on hard samples without sufficient time to leverage easy samples and build a solid foundation. Therefore, we set $h=0.85$ and multiply it with the metric function $D(\theta_{y_i})$, as will be further demonstrated in the experiments. $D(\theta_{y_i})$ can be formulated as: 

\begin{equation}
\begin{split}
    D(\theta_{y_i})= \sin\left(\frac{\theta_{y_i}}{2}\right), \theta_{y_i}\in[0,\pi],
\end{split}
\label{eq:sin}
\end{equation}
which monotonically increases with $\theta_{y_i}$, as shown in Fig. \ref{fg:function}b. This angle represents the distance of the sample from the positive class center, so that samples farther away from the positive class center are assigned a higher scale within the same training stage.

Our modulation coefficient function depends solely on the angle between the sample and the positive class center, while also acting as an adaptive adjust for both the positive and negative cosine similarities of hard samples. As the training progresses, $t^{(k)}$ increases, and after a certain point, the entire modulation coefficient function $\Phi>1$, which drives the model shifting its focus to hard samples, as illustrated in the blue curve of Fig. \ref{fg:mc2}. The angular margin $m$ for the hard sample is expanded, which indirectly reduces the positive cosine similarity of the hard sample, while directly enlarging the negative cosine similarity. By manipulating the loss in this manner, the model places greater emphasis on hard samples.

The specific process is illustrated in Fig. \ref{fg:flow}. After extracting the face features of the sample, we determine whether the sample is a hard sample. If the sample is a hard sample, its difficulty is assessed using Eq. \ref{eq:sin}, and then the modulation coefficient is calculated using Eq. \ref{eq:scale}. Finally, the modulation coefficient is applied to the angle margin $m$ of the sample's positive cosine similarity, and the negative cosine similarity is reweighted. let $P_{y_i}$ represent the probability of the $i$-th sample being correctly classified. The loss function formulation for AdaSin is as follows:
\begin{equation}
    L=-\frac{1}{N}\sum_{i=1}^{N}\log P_{y_i}
\label{eq:Adasinloss}
\end{equation}
where 
\begin{equation}
    P_{y_i}=\frac{e^{s\cos(\theta_{y_i}+\Phi\ast m)}}{e^{s\cos(\theta_{y_i}+\Phi \ast m)}+\sum_{j=1, j\neq y_i}^{n}e^{s(\Phi\ast\cos\theta_j)}}
\label{eq:AdaSinp}
\end{equation}

\subsection{Analysis}
In this section, we will analyse our designed modulation coefficient function. Let $t^{(k)}$ denote the positive cosine similarity of the $k$-th batch, and it has gone through $k$ iterations of EMA, as described by Eq. \ref{eq:ema}. For the convenience of our subsequent discussion, we expand the original equation from the first iteration of the EMA to the $k$-th iteration and $t^{(1)} = \alpha r^{(1)}$, as follows: 
\begin{equation}
\begin{split}
    t^{(k)} =&(1-\alpha)^0\alpha r^{(k)}+(1-\alpha)^1\alpha r^{(k-1)} \\
    &+ (1-\alpha)^2\alpha r^{(k-2)}+(1-\alpha)^3\alpha r^{(k-3)} \\
    &+...+(1-\alpha)^{k-1}\alpha r^{(1)},
\end{split}
\label{eq:texpand}
\end{equation}
then
\begin{equation}
    t^{(k)} =\alpha\sum_{j=1}^{k}(1-\alpha)^{k-j}r^{(j)}
    \label{eq:texpand2}
\end{equation}
In this equation, $k$ represents the total number of iterations, $\alpha$ is the update momentum, and $r^{(j)}$ is the average value of the positive cosine similarity for the current batch. The expression for $r^{(j)}$ in Eq. \ref{eq:texpand2} can be expanded as follows:
\begin{equation}
    r^{(j)}=\frac{1}{B}\sum_{i=1}^{B}\cos{\theta_{y_i}^{(j)}}
\end{equation}
Here, $\theta_{y_i}^{(j)}$ represents the angle between the $i$-th sample in the $j$-th batch and its corresponding positive class center, and $B$ is the number of samples in the batch. So the expression for $t^{(k)}$ can be rewritten as:
\begin{equation}
    t^{(k)}=\alpha\sum_{j=1}^{k}((1-\alpha)^{k-j}(\frac{1}{B}\sum_{i=1}^{B} \cos{\theta_{y_i}^{(j)}}))
\end{equation}

We assume that all samples in a batch are hard samples. The exist $n$ weight vectors corresponding to $n$ identity classes. Let $\theta_c$ be the angle from the decision boundary of the $c$-th class to its class center, where $1 < c < n $. Based on these assumptions, the angle range of a sample to the positive class center is $\theta_{y_i}\in(\theta_c, \pi)$. The range of the average positive cosine similarity for the samples in a batch is:

\begin{equation}
    -1<\frac{1}{B}\sum_{i=1}^{B}\cos{\theta_{y_i}^{(j)}}<\cos{\theta_c^{(j)}},
\end{equation}
then 
\begin{equation}
\begin{split}
    -\alpha\sum_{j=1}^{k}(1-\alpha)^{k-j}&<t^{(k)} \\
    & <\alpha\sum_{j=1}^{k}((1-\alpha)^{k-j}\cos{\theta_c^{(j)}}),
    \label{eq:trange}
\end{split}
\end{equation}
The range of values for our modulation coefficient function $D(\theta_{y_i})$ can also be expressed as:
\begin{equation}
    h\ast\sin{(\frac{\theta_c^{{(k)}}}{2})}<D(\theta_{y_i})<h
    \label{eq:sinrange}
\end{equation}

Since our modulation coefficient function $\Phi=t^{(k)}+D(\theta_{y_i})$ can be expanded as follows. 

\begin{equation}  
\begin{split}
     \Phi=\alpha\sum_{j=1}^{k}((1-\alpha)^{k-j}(\frac{1}{B}\sum_{i=1}^{B} \cos{\theta_{y_i}^{(j)}}))+D(\theta_{y_i})
\end{split}
\end{equation}
\label{eq:scale2}By combining Eq. \ref{eq:trange} and Eq. \ref{eq:sinrange}, the range of values for our modulation coefficient function can be expressed as:

\begin{equation}
\begin{split}
     -\alpha\sum_{j=1}^{k}(1-\alpha)^{k-j}+& h\ast\sin{(\frac{\theta_c^{{(k)}}}{2})} \\
     & <\Phi \\
     & <\alpha\sum_{j=1}^{k}((1-\alpha)^{k-j}\cos{\theta_c^{(j)}})+h
\end{split}
\label{eq:mcrange}
\end{equation}

\begin{figure}[h]
\centering
\includegraphics[width=0.7\linewidth]{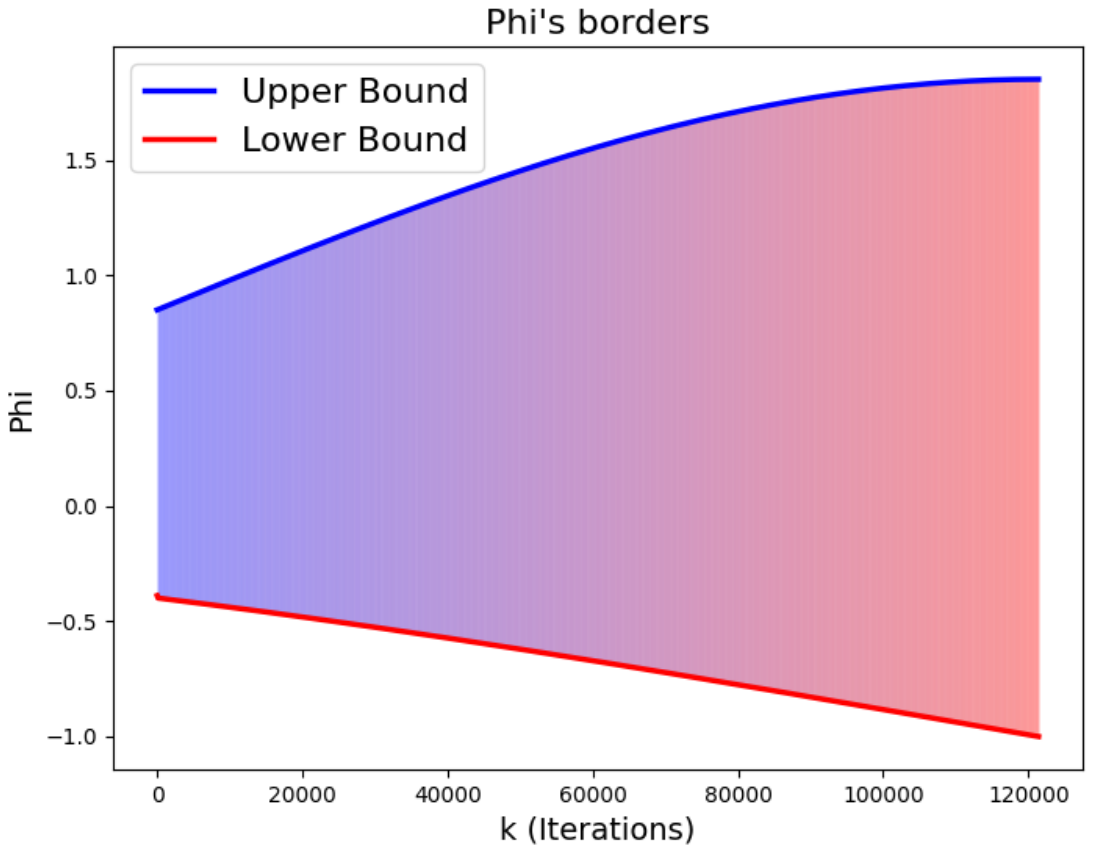}
\caption{Variable range of modulation coefficients function.}
\label{fg:PhiBoarders}
\end{figure}

Thus, the approximate range of our modulation coefficient function is illustrated in Fig. \ref{fg:PhiBoarders}. At the very beginning of the training, we fix $\theta_c$ at $\frac{\pi}{2}$, and as $k$ increases, $\theta_c$ gradually decreases until it reaches 0. It is apparent that in Eq. \ref{eq:mcrange} the lower bound of $\Phi$ only includes the decision boundary angle $\theta_c^{(k)}$ of the current batch $k$, and is not influenced by the decision boundary sizes from previous iteration batches. Furthermore, the constant term $\alpha\sum_{j=1}^{k}(1-\alpha)^{k-j}$ increases as the number of iterations increases, which also causes the lower bound of $\Phi$ to decrease over time, as shown in Fig. \ref{fg:PhiBoarders}. On the other hand, the cosine value $\cos{\theta_c^{(j)}}$ of the decision boundaries from previous iteration batches directly affects the upper bound of $\Phi$. This results in the gap between the upper and lower bounds of the modulation coefficient $\Phi$ growing larger.

\begin{figure}[h]
\centering
\includegraphics[width=0.7\linewidth]{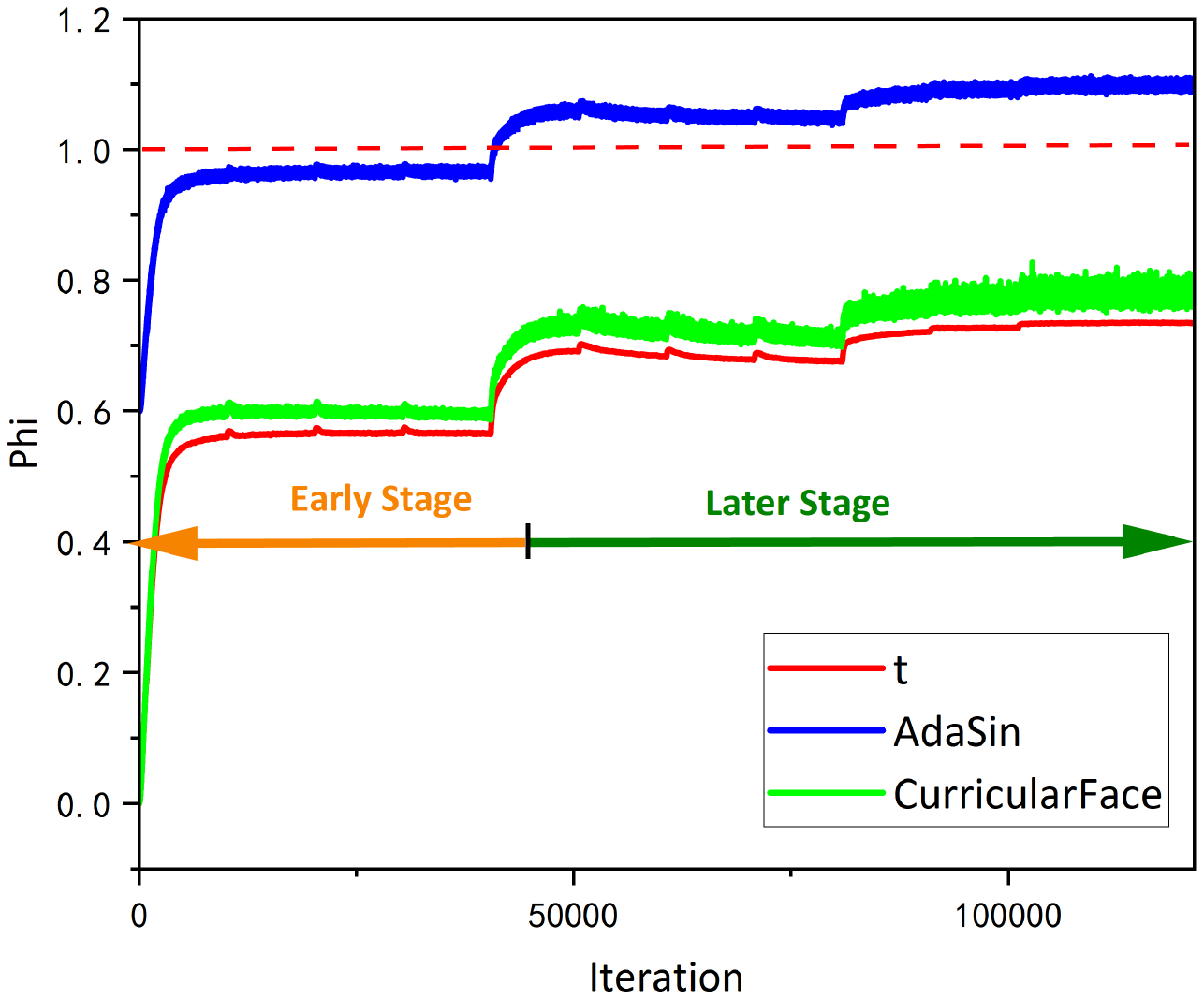}
\caption{The red solid line, green solid line, and blue solid line represent the trends of the adaptive parameter $t$, the modulation coefficients of CurricularFace, and the modulation coefficients of our AdaSin, respectively, throughout the training process. We can clearly observe that the modulation coefficient curve of CurricularFace (green solid line) remains consistently below the red dashed line throughout the entire training process.}
\label{fg:mc2}
\end{figure}

It follows that our modulation coefficient function is closely related to the variation of the decision boundary. We not only maintain the excellent mechanism of adaptive curriculum learning but also ensure both positive and negative cosine similarity of hard samples affected by the modulation coefficients, thus giving more importance to hard samples. As shown in Fig. \ref{fg:mc2}, our modulation coefficient exceeds 1 after 42k iterations, which means that after 42k iterations, the model starts to emphasize the hard samples.

\subsection{Optimization}
In this section, we will show that AdaSin can be easily optimized by the classical stochastic gradient descent (SGD) method.

Assuming $x_i$ is the embedding feature of the $i$-th sample and belongs to the $y_i$-th class, the input of the loss function is logits $f_j$, where $j$ denotes the $j$-th class. In the forwarding process, when $j=y_i$, $f_j$ has two cases: if the sample is a easy sample, then $f_j=s(\cos(\theta_{y_i}+m))$, otherwise $f_j=s(\cos( \theta_{y_i}+\Phi\ast m))$. When $j\neq y_i$, there are also two cases: $f_j=s(\cos\theta_j)$ if the sample is easy, otherwise $f_j=s(\Phi\ast\cos\theta_j)$. It is important to mention that the modulation coefficient function $\Phi$ has been constantized, so no gradient calculation will be performed on it. In the backward propagation process, the update gradients of $x_i$ and $w_i$ are also divided into four cases:

\begin{equation}
    \begin{split}
        \frac{\partial L}{\partial x_i} = 
        \begin{cases}
            \frac{\partial L}{\partial f_{y_i}}(s \frac{\sin{(\theta_{y_i} + m)}}{\sin{\theta_{y_i}}}) W_{y_i}, &j=y_i,easy \\
            \frac{\partial L}{\partial f_{y_i}}(s \frac{\sin{(\theta_{y_i} + \Phi \ast m)}}
            {\sin{\theta_{y_i}}})W_{y_i}, &j=y_i,hard \\
            \frac{\partial L}{\partial f_j}sW_j,& j \neq y_i, easy \\
            \frac{\partial L}{\partial f_j}s(\Phi)W_j,& j \neq y_i, hard
        \end{cases}
    \end{split}
\label{eq:gradientx}
\end{equation}

\begin{equation}
    \begin{split}
        \frac{\partial L}{\partial W_j} = 
        \begin{cases}
            \frac{\partial L}{\partial f_{y_i}}(s \frac{\sin{(\theta_{y_i} + m)}}{\sin{\theta_{y_i}}}) x_i, &j=y_i,easy \\
            \frac{\partial L}{\partial f_{y_i}}(s \frac{\sin{(\theta_{y_i} + \Phi \ast m)}}
            {\sin{\theta_{y_i}}})x_i, &j=y_i,hard \\
            \frac{\partial L}{\partial f_j}sx_i,& j \neq y_i, easy \\
            \frac{\partial L}{\partial f_j}s(\Phi)x_i,& j \neq y_i, hard
        \end{cases}
    \end{split}
\label{eq:gradientw}
\end{equation}

As can be seen from Eq. \ref{eq:gradientx} and Eq. \ref{eq:gradientw}, the gradient of Adasin's positive cosine similarity and negative cosine similarity for hard samples are both closely related to the modulation coefficient function $\Phi$. The entire training process is summarised in Algorithm \ref{ag:Adasin}.

\begin{algorithm}[H]
        \caption{: AdaSin}
        \label{ag:Adasin}
        \begin{algorithmic}[1]
	\Require Embedding features of the $i$-th sample $x_i$ and labels $y_i$ for the $i$-th sample. The weight vector $W$ of the class center, the angle $logits$ are the cosines of the feature vectors and class weight vectors, $m$ is the marginal value, and $K$ is the number of iterations required for an epoch.
            \For{$k = 1 \to K$} 
            \State Update the parameter $t$ by Eq. \ref{eq:ema}
            \State $\theta = \arccos(logits)$
            \State Compute the $\Phi $
            \If {$\cos{(\theta_{y_i} + m)} \geq \underset{0 \leq j \leq n-1}{\max} \cos \theta_j$}
            \State $ T(\theta_{y_i},m,\Phi) = \cos{(\theta_{y_i} + m)}$
            \Else
            \State $ T(\theta_{y_i},m,\Phi) = \cos{(\theta_{y_i} + \Phi\ast m)}$
            \EndIf
            \If {$ \cos{(\theta_{y_i} + m)} \geq \cos \theta_j$}
            \State $ N(\Phi,\theta_j) = \cos{\theta_j}$
            \Else 
            \State $ N(\Phi,\theta_j) =  \Phi \ast \cos{\theta_j}$
            \EndIf
            \State Compute the loss $L$ by Eq. \ref{eq:Adasinloss} and Eq. \ref{eq:AdaSinp}
            \State Compute the gradients of $x_i$ and $W$
            \State Update the parameters $x_i$ and $W$
            \EndFor 
            \Ensure $W, x_i$
        \end{algorithmic}
\end{algorithm}

\begin{table}
    \centering
    \setlength{\tabcolsep}{5mm}{
    \begin{tabular}{c|c}
        \hline
        \textbf{Methods} & \textbf{Decision Boundaries} \\
        \hline
         Softmax & $\cos{\theta_{y_i}}  = \cos{\theta_j}$   \\
        \hline
         ArcFace & $ \cos({\theta_{y_i} + m}) = \cos{\theta_j}$ \\
        \hline
         MV-Arc-Softmax & $\cos({\theta_{y_i} + m}) = \cos{\theta_j} $ (easy) \\ 
                        & $\cos({\theta_{y_i} + m})  = (t+1)\cos{\theta_j} + t$ (hard)\\
        \hline
         CurricularFace & $\cos({\theta_{y_i} + m}) = \cos{\theta_j} $ (easy)  \\
                        & $\cos({\theta_{y_i} + m}) = (t^{(k)}+ \cos{\theta_j})\cos{\theta_j}$ (hard)\\
        \hline
         AdaSin(Ours)   & $\cos({\theta_{y_i} + m})  = \cos{\theta_j} $ (easy) \\
                        & $\cos({\theta_{y_i} +\Phi \ast m}) = \Phi \ast \cos{\theta_j} $ (hard)\\
        \hline
    \end{tabular}
    }
    \caption{Decision boundaries for different methods}
    \label{tb:Db}
\end{table}

\begin{figure}
\centering
\includegraphics[width=0.7\linewidth]{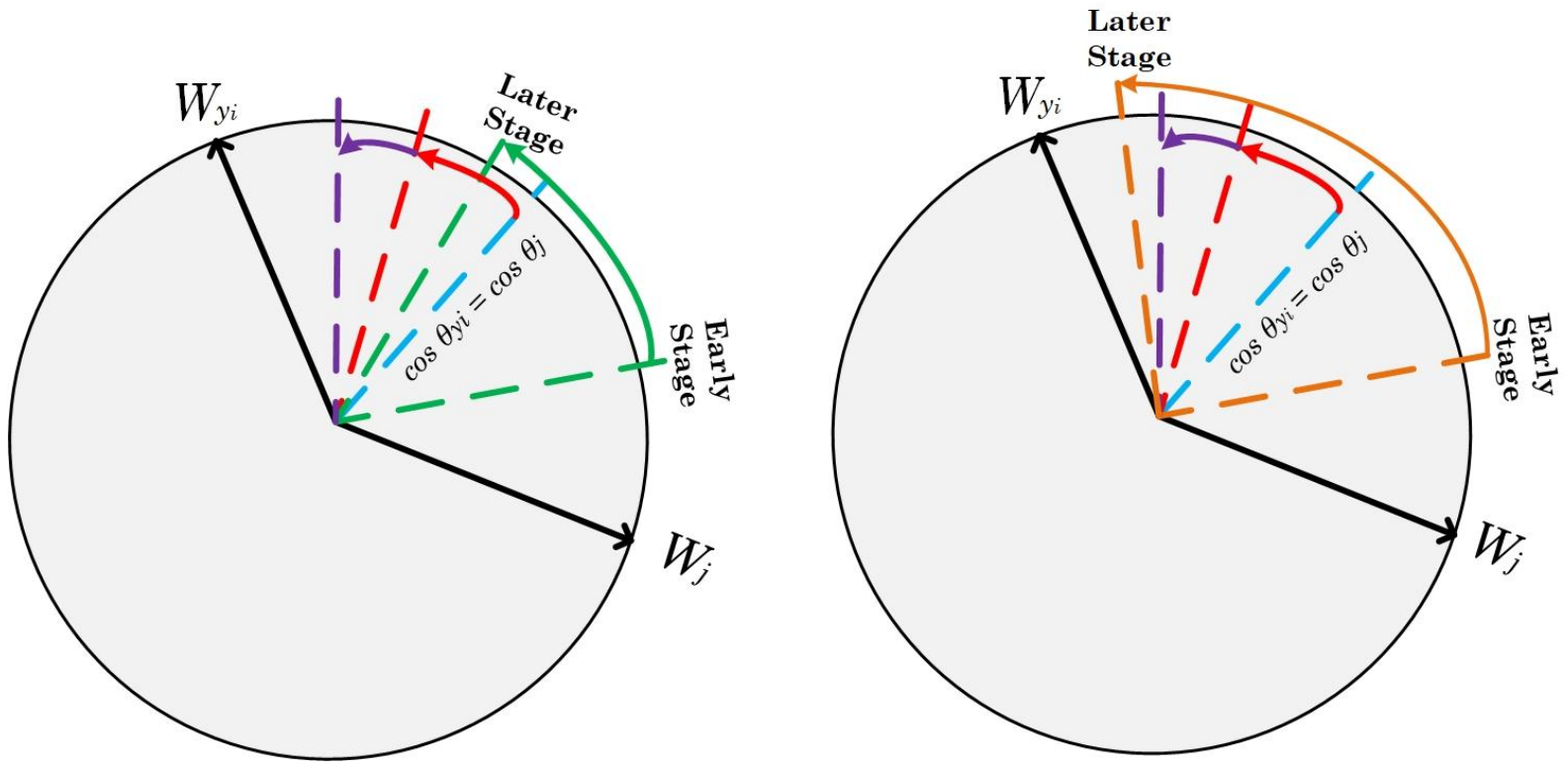}
\caption{$W_{y_i}$ and $W_j$ denote positive class centers and negative class centers. The \textcolor{blue}{blue dashed line}, \textcolor{red}{red dashed line}, \textcolor{violet}{violet dashed line}, \textcolor{green}{green dashed line} and \textcolor{orange}{orange dashed line} denote the decision boundaries of traditional softmax, ArcFace, MV-Arc-Softmax, CurricularFace and AdaSin respectively, and the arrows indicate their variations.}
\label{fg:Db}
\end{figure}

The decision boundary of our AdaSin is more flexible and spans a wider range of decision boundaries compared to those proposed in previous methods. In Table. \ref{tb:Db}, the decision boundaries of previous methods and our method are listed. When the samples are hard samples, our modulation coefficient not only adaptively adjusts the negative cosine similarity, but also adaptively adjusts the angle margin of the positive cosine similarity. This results in more pronounced changes in the decision boundaries. In Fig. \ref{fg:Db}, the green dashed line on the left represents the decision boundary of CurricularFace. It can be seen that in the early stages of training, the decision boundary is positioned near the right end of $cos{\theta_{y_i}}=cos{\theta_j}$, close to the center of the negative class. This is due to the modulation coefficient being less than 1 in the early stages of training. During this phase, the model allows a larger feasible region for hard samples. As the modulation coefficient increases, the decision boundary moves closer to the positive class center. However, even though the decision boundary shifts to the left of $cos{\theta_{y_i}}=cos{\theta_j}$ in the later stages, the modulation coefficient remains less than 1, keeping the decision boundary still far from the positive class center. In contrary, with AdaSin, the modulation coefficient is greater than 1 in the late stage of training, pulling the decision boundary closer to the positive class center and significantly reducing the feasible area for hard samples.


\begin{figure}
\centering
\includegraphics[width=0.7\linewidth]{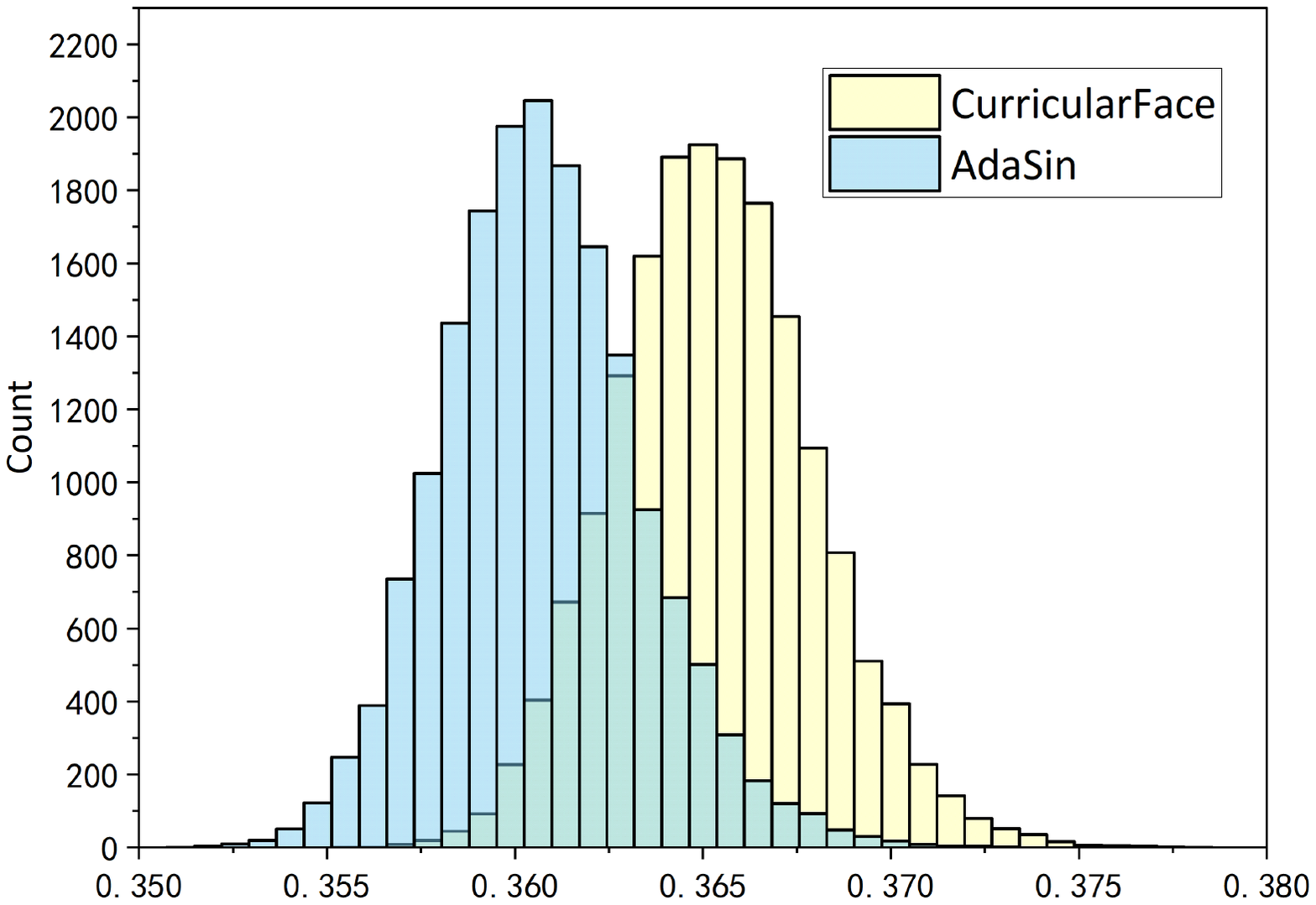}
\caption{Comparison of the $D(\theta_{y_i})$ values of samples in each iteration during the last two epochs between AdaSin and CurricularFace.}
\label{fg:angle}
\end{figure}

Fig. \ref{fg:angle} shows the mean sine value of the angle between the feature vectors extracted by our method and CurricularFace in each iteration during the last two epochs, relative to the positive class center, which we denote as $D(\theta_{y_i})$. It can be clearly observed that our method brings the samples closer to the class center, meaning that AdaSin is more effective at increasing intra-class compactness. The vertical axis represents the number of iterations.

\section{Experiments}
\label{sc:ex}

\subsection{Experimental Settings}

We cropped $112 \times 112$ face images using five landmarks \citep{zhang2016joint,tai2019towards}, following the method used in CurricularFace \citep{huang2020curricularface}. For fairness, all experiments were conducted using IResNet50 \citep{duta2021improved} as the backbone network. The entire framework was implemented using Pytorch \citep{paszke2017automatic}. The model was trained on two NVIDIA GeForce RTX 4090 GPUs with a batch size of 512. We used the SGD optimizer with a momentum of 0.9 and weight decay of 5e-4. The learning rate was initialized at 0.1 and reduced by a factor of 10 at the 4th, 8th, and 10th epochs, for a total of 12 epochs. The hyperparameters were set according to the settings of ArcFace \citep{deng2019arcface}, with $s = 64$ and $m = 0.5$. 

\subsection{Datasets}

We used the MS1MV3 \citep{deng2019lightweight} dataset for training, applying random horizontal flipping to the images. The MS1MV3 dataset, a widely used benchmark in face recognition, is an enhanced version of the original MS-Celeb-1M\citep{guo2016ms}. It improves upon the original by removing mislabeled data and focusing on real-world variations such as pose and illumination, thereby refining the quality and diversity of the labeled images. The dataset contains 5.17 million face images from over 93,000 unique individuals and serves as one of the key benchmark datasets in the field of face recognition. 

We extensively evaluated our method on several well-known benchmarks, including LFW. \citep{huang2008labeled}, CALFW \citep{zheng2017cross}, CPLFW \citep{zheng2018cross}, AgeDB-30 \citep{moschoglou2017agedb}, CFP-FP \citep{sengupta2016frontal}, VGG2-FP \citep{cao2018vggface2}, IJB-B \citep{whitelam2017iarpa}, and IJB-C \citep{maze2018iarpa}. LFW contains over 13,000 real-world face images for unconstrained face verification. Additionally, we evaluate the model's generalization ability under age variations using CALFW and AgeDB-30, and its adaptability to pose variations using CPLFW, CFP-FP, and VGG2-FP, which focus on extreme pose changes in face recognition. IJB-B is an early version in the IJB series, containing 21,798 images, 5,712 video frames, and 1,845 unique identities. This dataset features a diverse range of poses, lighting conditions, and resolutions. For the 1:1 verification task, it includes 10,270 positive sample pairs and 8,000,000 negative sample pairs. IJB-C includes 31,334 static images and 11,779 video frames, involving 3,531 independent identities. In the 1:1 verification task, IJB-C has 19,557 positive sample pairs and 15,638,932 negative sample pairs.

\subsection{Effect of Hyperparameter $h$}
In this section, we explore the impact of different $h$ values on AdaSin's performance. To ensure the model emphasizes hard samples in later training stages, this section explores the effect of different $h$ values on AdaSin's performance. Table. \ref{tb:hexp} lists the possible range of values we consider for $h \in [0.8,1]$ and compares face verification results for AdaSin across different $h$ values on IJB-B (TAR at FAR = $1\times10^{-5}$ and FAR = $1\times10^{-4}$) as well as CPLFW, AgeDB-30, and CFP-FP. AdaSin is highly sensitive to $h$, when $h=1$, the metric function $D(\theta_{y_i})$ remains unchanged. When $h = 0.85$, the modulation coefficient ensures that the model has sufficient time in the early stages to leverage easy samples and solidify its foundation, and the performance of AdaSin approaches saturation.

\begin{table}[h]
    \centering
    \resizebox{\linewidth}{!}{
    \begin{tabular}{c|ccc|cc}
    \toprule[1pt]
        \hline
        \makecell*[c]{\textbf{Methods}} & \textbf{CPLFW} & \textbf{Agedb} & \textbf{CFP-FP} & \multicolumn{2}{c}{\textbf{IJB-B}} \\ \cline{5-6} 
        & & & & FAR = 1e-5 & FAR = 1e-4 \\ \hline
        $h = 1.0$ & 92.62 & 98.00 & 97.84 & 90.68 & 95.04 \\
        $h = 0.95$ & 92.63 & 98.02 & 98.13 & 91.32 & 95.04 \\
        $h = 0.90$ & 92.47 & 98.02 & 97.97 & 90.49 & 94.97\\
        $h = 0.85$ & \textbf{92.67} & 98.12 & \textbf{98.13} & \textbf{91.69} & \textbf{94.98} \\
        $h = 0.80 $ & 92.55 & \textbf{98.15} & 98.08 & 90.86 & 94.95 \\
        \hline
    \end{tabular}
    }
    \caption{Performance validation on CPLFW, AgeDB-30, CFP-FP, and IJB-B (TAR at FAR = $ 1 \times 10^{-5}$ and FAR = $1\times 10^{-4}$) for different $h$ values.}
    \label{tb:hexp}
\end{table}

\subsection{Ablation Experiments}
In this section, we explore the impact of dual adaptive penalty. Previous research has shown that adaptive margin penalties are generally more effective than fixed margins, and our dual adaptive penalty further enhances performance. First, we apply the modulation coefficient function $\Phi$ to reweight the negative cosine similarity of ArcFace, naming it AdaSin-N. Next, we apply the modulation coefficient $\Phi$ to scale the angle margin $m$ in the positive cosine similarity of ArcFace, naming this variant AdaSin-T. Finally, we simultaneously reweight the negative cosine similarity and scale the angle margin in the positive cosine similarity using the modulation coefficient $\Phi$, naming it AdaSin-D. Verification is conducted on IJB-C (TAR at FAR = $1\times10^{-6}$ and FAR = $1\times10^{-5}$) and CPLFW, AgeDB-30, and CFP-FP datasets. Table. \ref{tb:TND} show that the model performs best when both cosine similarities incorporate the adaptive modulation coefficient function. 

\begin{table}[H]
    \centering
    \resizebox{\linewidth}{!}{
    \begin{tabular}{c|ccc|cc}
    \toprule[1pt]
        \hline
        \makecell*[c]{\textbf{Methods}} & \textbf{CPLFW} & \textbf{Agedb} & \textbf{CFP-FP} & \multicolumn{2}{c}{\textbf{IJB-C}} \\ \cline{5-6} 
        & & & & FAR = 1e-6 & FAR = 1e-5 \\ \hline
        ArcFace & 92.40 & 98.10 & 98.03 & 90.67 & 94.74 \\
        AdaSin-T & 92.62 & 98.02 & 98.03 & 90.00 & 94.87 \\
        AdaSin-N & 92.50 & \textbf{98.15} & 98.06 & 90.55 & 94.74 \\
        AdaSin-D & \textbf{92.67} & 98.12 & \textbf{98.13} & \textbf{91.37} & \textbf{94.90}\\
        \hline
    \end{tabular}
    }
    \caption{Performance evaluation metrics for different methods.}
    \label{tb:TND}
\end{table}

\begin{figure}[h]
    \centering
    \includegraphics[width=0.7\linewidth]{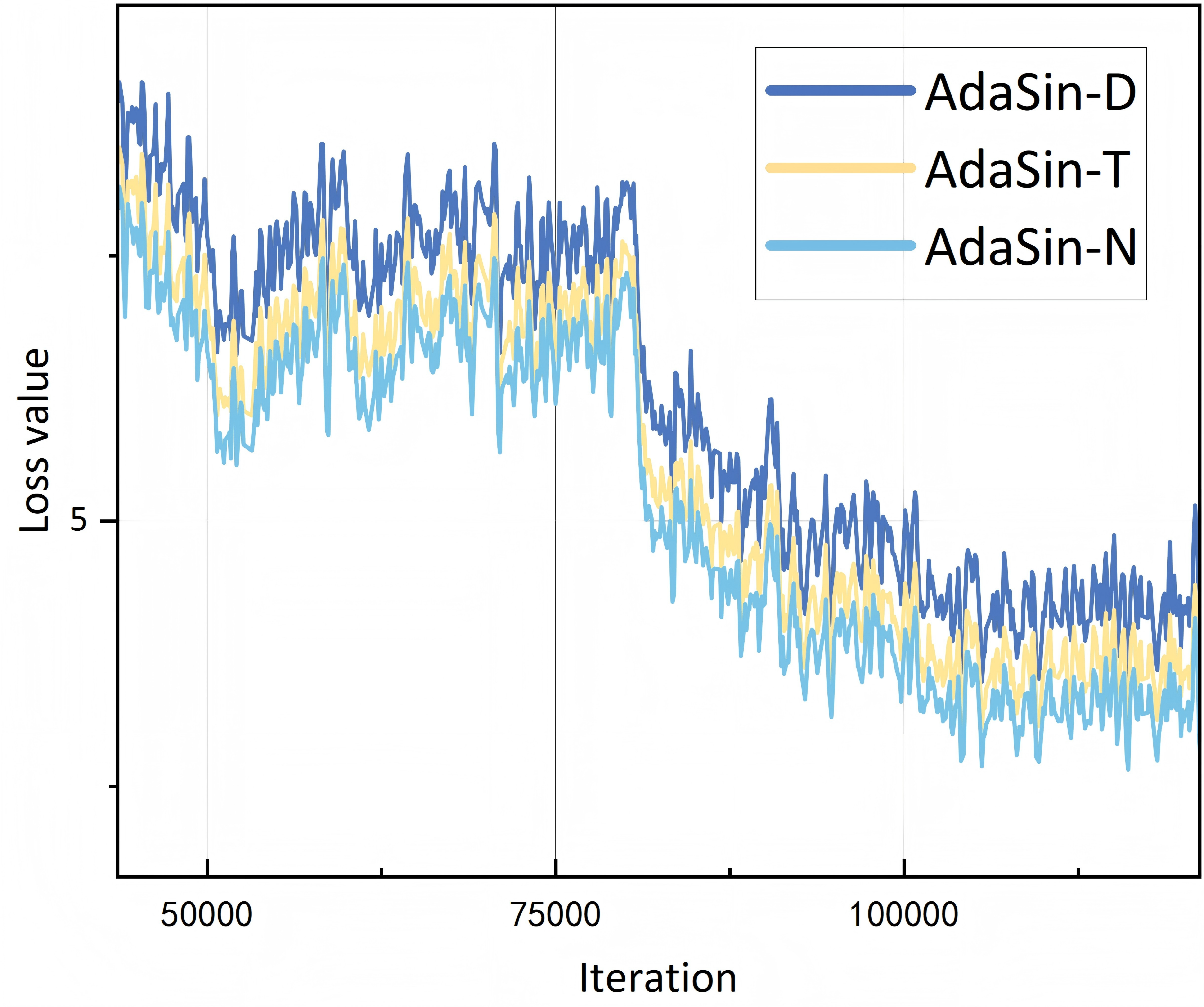}
    \caption{Illustration of the loss curves of the single-adaptive and dual-adaptive mechanisms in the late stages of training.}
    \label{fig:DTN}
\end{figure}

In this ablation study, Fig. \ref{fig:DTN} shows the loss curves for each of the three methods, recorded every hundred iterations. AdaSin-D, which incorporates a dual adaptive penalty, exhibits a higher loss value in the later stages of training compared to AdaSin-T and AdaSin-N, which use only a single adaptive mechanism. This is because the dual adaptive penalty applies a stronger penalty to hard samples, causing the model to place more emphasis on these challenging samples as training progresses.

\subsection{Convergence of Training }

\begin{figure}[H]
    \centering
    \includegraphics[width=0.7\linewidth]{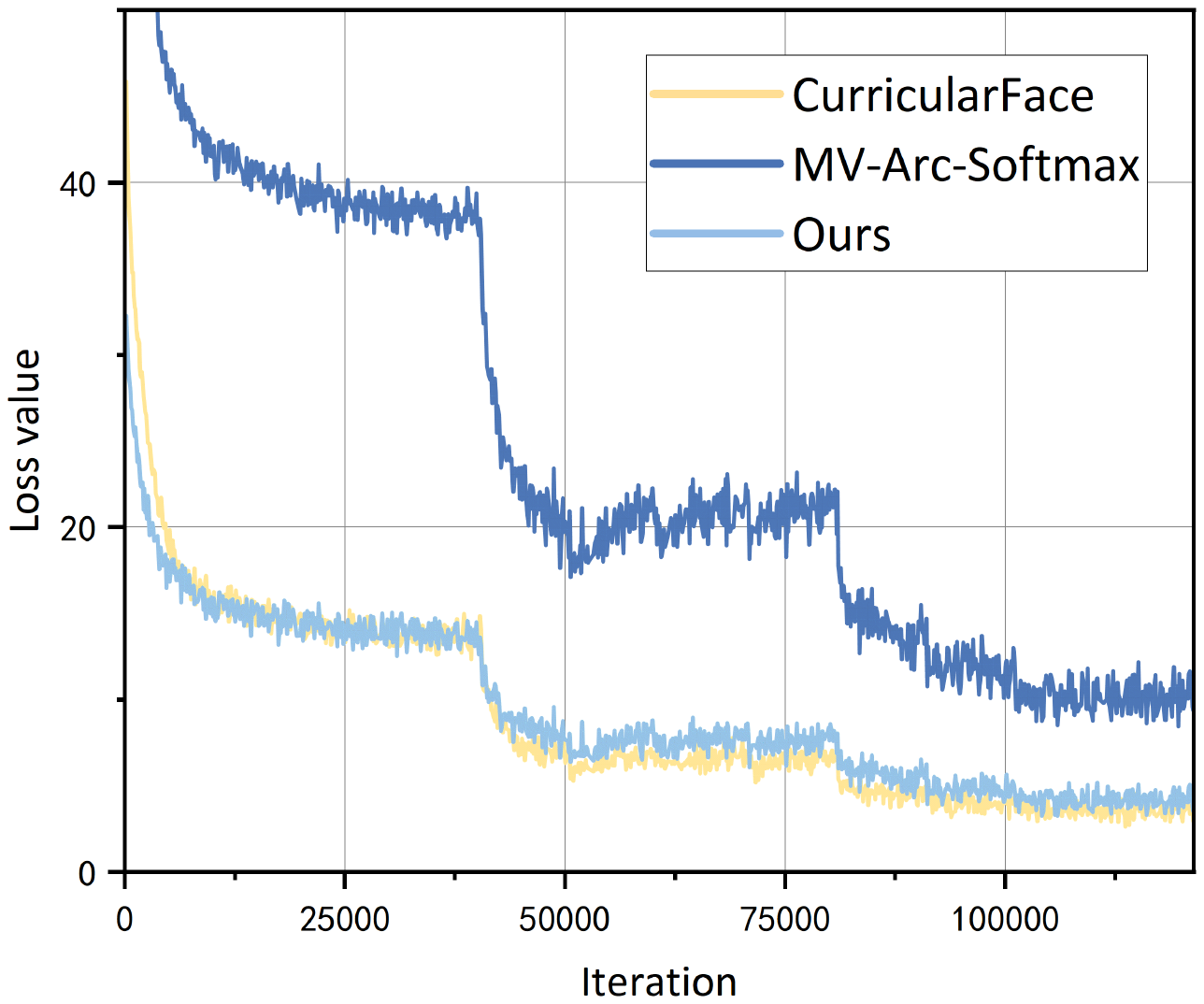}
    \caption{Illustration of loss curves per 100 iterations of adaSin, curricularFace, and mv-arc-softmax trained with backbone IResNet50.}
    \label{fig:loss}
\end{figure}

In this subsection, we analyze the changes in the AdaSin loss curve. We use IResNet50 as the network architecture and train it on the MS1MV3 dataset. As shown in Fig. \ref{fig:loss}, the light blue, yellow, and dark blue curves represent AdaSin, CurricularFace, and MV-Arc-Softmax, respectively. Compared to CurricularFace, AdaSin exhibits a lower loss value in the early stages of training, particularly before 6,000 iterations. This is because the initial phase includes a significant number of hard samples, and the dual adaptive penalty reduces the penalty on hard samples while prioritizing easier ones, resulting in a lower loss value. After the 6,000th iteration, AdaSin's loss value becomes comparable to that of CurricularFace, as the model stabilizes with a focus on easier samples. As the training progresses into the later stages, AdaSin's loss value becomes slightly higher than that of CurricularFace. This occurs because our modulation coefficient exceeds 1 at this point, leading to greater penalties for hard samples and placing more emphasis on them. Toward the end of training, especially after 100,000 iterations, the loss values for AdaSin and CurricularFace converge and stabilize. However, the loss value of MV-Arc-Softmax fluctuates throughout training. This instability is attributed to MV-Arc-Softmax's lack of an adaptive mechanism that adjusts based on the difficulty of hard samples.

\subsection{Comparisons with SOTA Menthods}

For fairness, we trained AdaSin along with several other methods on the MS1MV3 dataset using IResNet50 with identical parameter settings, and compared their performance against SOTA competitors across multiple benchmarks.

\subsubsection{Results on LFW, CALFW, CPLFE, AgeDB-30, CFP-FP and VGG2-FP}

As shown in Table. \ref{tb:accuracy}, AdaSin achieves comparable results to other competitors on the LFW benchmark, where performance is nearly saturated. It ranks first on the AgeDB-30 and CPLFW benchmarks. Moreover, CurricularFace performs less effectively than MV-Arc-Softmax and ArcFace on these benchmarks, confirming the limitations of CurricularFace as discussed earlier. It can be observed that AdaSin performs suboptimally on CALFW, CFP-FP, and VGG2-FP. This is because AdaSin tends to focus more on hard samples, which results in insufficient optimization of some easier samples. As a result, the overall performance of the model is limited on these specific benchmarks. We plan to improve AdaSin's performance on these benchmarks in future research.

\begin{table}[h]
    \centering
    \resizebox{\linewidth}{!}{
    \label{tb:accuracy}
    \begin{tabular}{c|cccccc}
        \toprule[1pt]
        \hline
        \textbf{Methods(\%)}& \textbf{LFW} & \textbf{CALFW} & \textbf{AgeDB-30} & \textbf{CPLFW} & \textbf{CFP-FP} & \textbf{VGG2-FP} \\
        \hline
        ArcFace & 99.82 & \textbf{96.12} & 98.10 & 92.45 & 98.03 & 95.22 \\
        CosFace & 99.82 & 96.08 & 98.05 & 92.55 & 97.93 & 95.34 \\
        Dyn-ArcFace & \textbf{99.83} & 96.05 & 98.00 & 92.60 & 97.93 & 95.46 \\
        ElasticFace-Arc+ & 99.80 & 96.08 & 98.05 & 92.37 & 98.07 & 95.28 \\
        AdaFace & 99.82 & 95.98 & 98.07 & 92.55 & \textbf{98.21} & \textbf{95.64}  \\
        MV-Arc-Softmax & 99.80 & 95.98 & 97.95 & 92.60 & 97.91 & 95.54  \\
        CurricularFace & 99.78 & 95.97 & 97.92 & 92.20 & 97.30 & 95.00 \\
        \hline
        AdaSin(Ours) & 99.82 & 96.02 &\textbf{98.12}& \textbf{92.67} & 98.13 & 95.18 \\
        \hline
    \end{tabular}
    }
    \caption{Comparison of performance with other SOTA methods on benchmarks including LFW, CALFW, CPLFW, AgeDB-30, CFP-FP, VGG2-FP.}
\end{table}

\subsubsection{Results on IJB-B and IJB-C}
\begin{table}[h]
    \centering
    \setlength{\tabcolsep}{5mm}{
    \begin{tabular}{c|cc}
        \toprule[1pt]
        \hline
        \textbf{Methods(\%)} &\textbf{IJB-B} & \textbf{IJB-C}  \\
        \hline
        ArcFace & 91.09 & 94.74\\
        CosFace & 90.51 & 94.63\\
        Dyn-ArcFace & 90.90 & 94.69\\
        ElasticFace-Arc+ & 90.73 & 94.78\\
        AdaFace & 91.31 & 94.86 \\
        MV-Arc-Softmax & 91.02 & 94.86 \\
        CurricularFace & 90.02 & 94.41\\
        \hline
        AdaSin(Ours) & \textbf{91.69} & \textbf{94.90} \\
        \hline
    \end{tabular}
    }
    \caption{Comparison of performance with other SOTA methods on IJB-B and IJB-C (TAR at FAR = 1e-5). }
    \label{tb:IJB}
\end{table}

Table. \ref{tb:IJB} shows the performance of different methods on IJB-B and IJB-C. In the 1:1 verification task, our AdaSin outperforms the SOTA methods once again. When comparing the TAR at FAR = 1e-5 for AdaSin on IJB-B and IJB-C with other SOTA methods, AdaSin achieves the best performance. This is a demonstration of AdaSin's superiority in complex datasets with a large number of hard samples.

\section{Conclusion}

In this paper, we propose an improved modulation coefficient function by adopting a novel method to measure sample difficulty. Additionally, we design a loss function with a dual-adaptive margin penalty mechanism, which is compatible with the concept of curriculum learning. Our loss function is capable of assigning different penalties to samples with varying levels of difficulty, making the features within the same class more compact and enhancing the discriminability of features across different classes. This enables the model to extract more discriminative face features from hard-to-recognize samples. AdaSin achieves outstanding results on several challenging benchmarks, demonstrating superior performance compared to other SOTA competitors. In future research, we will aim to challenge AdaSin to improve its performance in specific environments.

\section*{Acknowledgment}
This work was supported in part by the Guangdong Basic and Applied Basic Research Foundation under Grant 2022A515110020, the Jinan University "Da Xiansheng" Training Program under Grant YDXS2409, the National Natural Science Foundation of China under Grant 62271232, and the National Natural Science Foundation of China under Grant 62106085.




\end{document}